\documentclass[letterpaper, 10 pt, conference]{ieeeconf}
\IEEEoverridecommandlockouts                              

\usepackage{graphicx} 
\usepackage{amsmath} 
\usepackage{amssymb}  
\usepackage{float}
\usepackage[colorinlistoftodos, textwidth=4cm, shadow]{todonotes}
\usepackage[ruled]{algorithm2e}

\usepackage{url}

\usepackage{tikz}
\usepackage{textcomp}
\usepackage{hyperref}
\usepackage{lipsum}

\def\doi{DOI: \href{https://doi.org/XX.YYYYY/X.YYYYY}{XX.YYYYY/X.YYYYY}}
\newcommand\copyrighttext{%
  \footnotesize \textcopyright \the\year{} IEEE. Personal use of this material is permitted.
  Permission from IEEE must be obtained for all other uses, in any current or future
  media, including reprinting/republishing this material for advertising or promotional
  purposes, creating new collective works, for resale or redistribution to servers or
  lists, or reuse of any copyrighted component of this work in other works.
  }
\newcommand\copyrightnotice{%
\begin{tikzpicture}[remember picture,overlay]
\node[anchor=south,yshift=10pt] at (current page.south) {\fbox{\parbox{\dimexpr\textwidth-\fboxsep-\fboxrule\relax}{\copyrighttext}}};
\end{tikzpicture}%
}

\newcommand{\bm}[1] {\ensuremath \boldsymbol{#1} }
\newcommand{\argmax}{\operatornamewithlimits{argmax}}

\title{\LARGE \bf
Streaming Scene Maps for Co-Robotic Exploration in Bandwidth Limited Environments
}

\author{Yogesh Girdhar$^{1}$, Levi Cai$^{2}$, Stewart Jamieson$^{3}$, Nathan McGuire$^{4}$, Genevieve Flaspohler$^{3}$,\\ Stefano Suman$^{1}$, and Brian Claus$^{1}$
\thanks{*This work was supported by NSF-NRI Award Number 1734400}
\thanks{$^{1}$ Y. Girdhar, S. Suman, and B. Claus are with the Applied Ocean Physics and Engineering Department at the Woods Hole Oceanographic Institution (WHOI)
        {\tt\small \{yogi, ssuman, bclaus\}@whoi.edu}}%
\thanks{$^{2}$L. Cai is with the MIT Media Lab, but performed this work as a guest student at WHOI
        {\tt\small cail@mit.edu }}%
\thanks{$^{3}$S. Jamieson and G. Flaspohler are with the MIT-WHOI Joint Program in Applied Ocean Science and Engineering
        {\tt\small \{sjamieson, gflaspohler\}@whoi.edu }}%
\thanks{$^{4}$N. McGuire is with the Mechanical Engineering Department at Northeastern University, but performed this work as a guest student at WHOI
        {\tt\small mcguire.n@husky.neu.edu }}%
}

\begin{document}

\maketitle
\copyrightnotice
\thispagestyle{empty}
\pagestyle{empty}

\begin{abstract}
This paper proposes a bandwidth tunable technique for real-time probabilistic scene modeling and mapping to enable co-robotic exploration in communication constrained environments such as the deep sea. The parameters of the system enable the user to  characterize the scene complexity represented by the map, which in turn determines the bandwidth requirements. The approach is demonstrated using an underwater robot that learns an unsupervised scene model of the environment and then uses this scene model to communicate the spatial distribution of various high-level semantic scene constructs to a human operator. Preliminary experiments in an artificially constructed tank environment as well as simulated missions over a 10m$\times$10m coral reef using real data show the tunability of the maps to different bandwidth constraints and science interests. To our knowledge this is the first paper to quantify how the free parameters of the unsupervised scene model impact both the scientific utility of and bandwidth required to communicate the resulting scene model.


\end{abstract}

\section{Introduction}

The challenges of exploration in remote and extreme environments such as the deep seas \cite{barker2016, cressey2015ocean}, cave systems \cite{tardioli2015}, outer space \cite{gao2017} and during or after a natural disaster \cite{nagatani2013,yuan2017aerial} have much in common. It is expensive and inherently dangerous for humans to explore such locations directly; hence, the use of mobile robots is desirable. However, if communication bottlenecks exist in the environment, prohibiting live streaming of video or other sensor data, then direct control of the robots is generally not possible. This paper describes a novel approach to co-robotic exploration in communication starved environments, and presents a system implementation of an under-sea exploration robot for co-robotic exploration of marine environments.

Although physically controlling a robot can be achieved over relatively low bandwidth, it is difficult to transmit the scene information necessary for an operator or scientist to make high level navigational decisions. We propose a spatially correlated Chinese Restaurant Process (CRP)-based \cite{Teh2010} scene understanding model, that can be tuned to operate with the available bandwidth, for characterizing the environment in a manner useful for describing the operator's scientific interest. 

The most common approach to underwater exploration today is to either use a tethered vehicle or an autonomous underwater vehicle (AUV) to traverse a pre-planned path while collecting sensor data, which is reviewed once the robot returns to a location (the surface or a docking station) where high speed communication is possible \cite{Singheaan4809}. Such missions are useful for collecting population statistics for dense and stationary phenomena, but have limited utility when collecting data on spatially patchy or transient phenomena, especially in new or poorly mapped locations. 
\begin{figure}[t]
    \centering
    \includegraphics[width=1\linewidth]{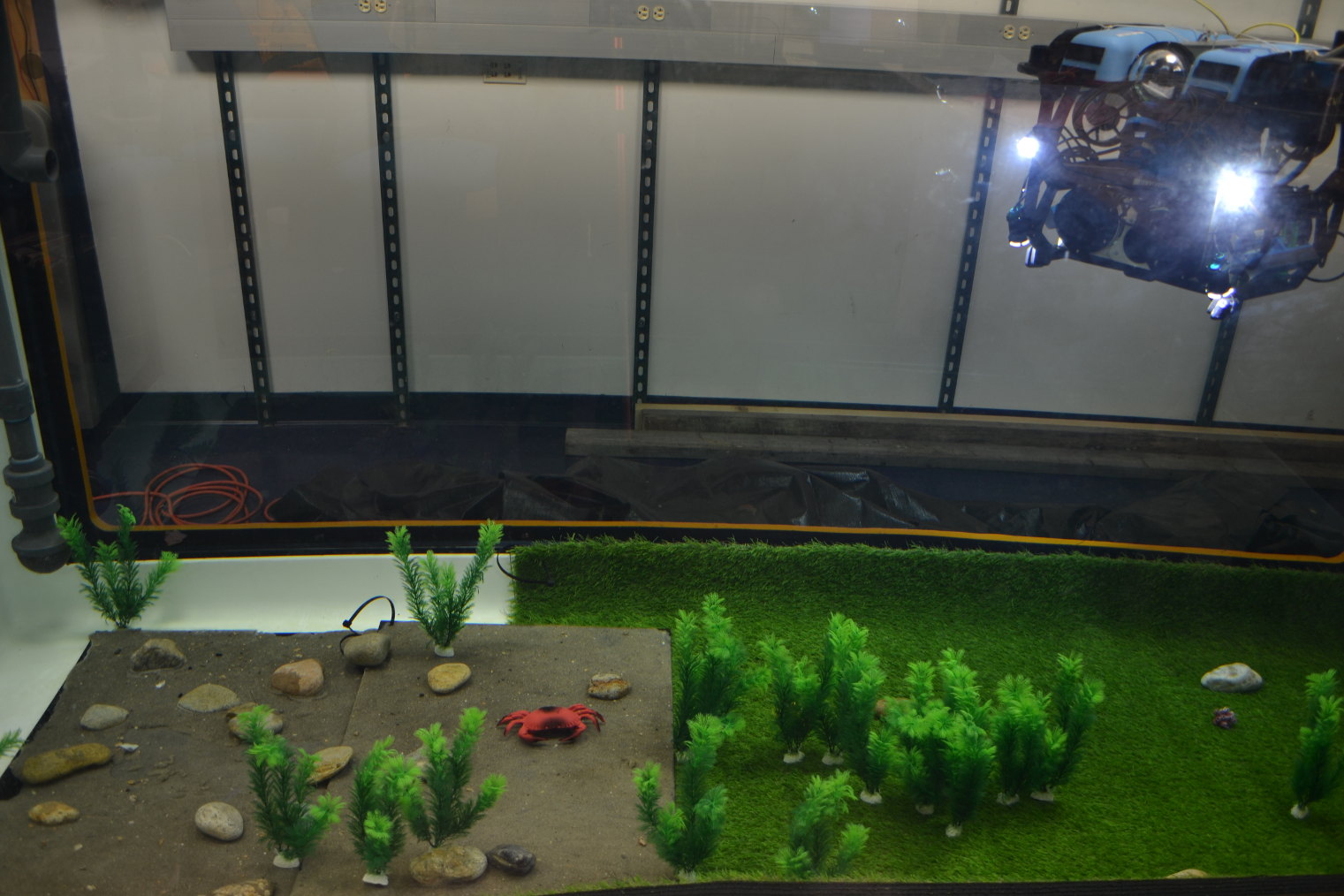}
    \caption{The proposed co-robotic exploration system can be used to adaptively collect data based on visual information, while keeping a scientist in the loop. Here we show an AUV developed in our lab, exploring an artificially created underwater environment.}
\label{fig:tankrobot}
\end{figure}

\section{Background}




Scientific robots have been deployed to study a variety of extreme environments, necessitating flexible models that represent scientifically relevant detail and structure in the environment.  Simultaneous localization and mapping (SLAM) techniques have been used to address the challenge of robot localization and to construct occupancy and geometric maps of terrestrial caves and mines \cite{Thrun2004, weidner2017underwater}, as well as in marine environments \cite{kinsey2006survey}. However, the occupancy maps used in SLAM are often not sufficient for modeling complex scientific phenomenon.

Gaussian processes (GPs) \cite{Rasmussen:2006:GP} and Gaussian mixture models have seen widespread application for modeling smooth, continuous environmental phenomena, such as chemical plumes \cite{Jakuba2011} and plankton concentrations \cite{Das2015}, in autonomous science applications. 
Other work has sought to explicitly model discrete, causal structure in scientific environments using Bayesian Networks (BN). Bayesian network models are suitable when generative model of the phenomena to be explored is known, but has its own set of challenges. Exact inference in BN can be difficult in all but very small problems. Arora et al. \cite{Arora2017clean} have used BN's to model geological phenomena; their formulation relies on a significant amount of prior domain knowledge specified by scientific experts in the form of conditional probability tables. Other works have used fixed thresholds on scalar sensor data set manually by domain scientists to dictate autonomous behavior \cite{Zhang2016} or rely on coarse prior maps of the environment, which are available when planning a mission \cite{Gautam2017}. While relying on prior scientific expertise or survey data is sometimes realistic, in environments such as the deep sea or space domain scientists may not have enough \textit{a priori} knowledge to formulate a scientific model and hypotheses. The unsupervised terrain model used in this work does not require prior scientific expertise, training data, or environmental maps; instead, a terrain model is learned directly from discrete visual word observations in a camera stream.

Bayesian nonparametric models have been used extensively for unsupervised scene understanding due to their ability to characterize complex scenes with unknown complexity in terms of number of objects. Furthermore, strong Bayesian priors on the scene structure can enable realtime in-situ learning, even with high dimensional sensor data streams \cite{Girdhar2015Gibbs}. Sudderth et al. \cite{Sudderth2009} have proposed the use of thresholded global positioning system (GPS) for modeling spatial distribution of scene constructs. Joho et al. \cite{Joho2012} have used a Dirichlet Process in combination with a Beta process to represent indoor scene structure in terms of objects and their spatial configuration. Steinberg et al. \cite{steinberg2011bayesian} employ a Dirichlet process clustering technique to learning underwater terrain models.  In our prior work \cite{Girdhar2015} we proposed heirarchical Dirichlet process realtime online spatiotemporal topics (HDP-ROST), which uses a spatially coupled CRP to model the spatial distribution of observed substrate types. This unsupervised model allows for structure learning directly from visual observations, without the need for prior scientific knowledge about an environment or smoothness constraints. This paper combines HDP-ROST based scene understanding approach with a physical platform for bandwidth-limited robot autonomy to present a cohesive system for human-robot co-robotic exploration of marine environments.


From a hardware perspective, platforms for co-robotic exploration often seek to overcome the challenges of extreme environments using innovative physical designs. Ocean exploration using a tethered Remotely Operated Vehicles (ROV) is perhaps the most common approach to co-robotic scientific exploration \cite{michel2018,mcveigh2018,everett2018}, as it enables high speed real-time relay of all sensor data to a remote scientist, and enable low-level control of the vehicle.  However some ocean environments such as under the Arctic ice have unique challenges that prevent the deployment of a regular ROV. The moving ice can make it difficult for a ship to keep station and deploy ROVs. The Nereid Under Ice \cite{jakuba2018} vehicle uses a novel approach to co-robotic exploration under-ice by maintaining a high-speed communication link through a hair-thin fiber-topic cable. This cable is unspooled from both the ship and the robot, and the robot maintains the capability to return to the open ocean to be recovered in case the tether breaks.

Conversely, long duration ocean observing platforms such as underwater gliders relay subsets of water column data over low speed satellite links when at the surface to direct future sampling efforts \cite{mensi2014,flexas2018}.  Remotely gathered data is also fed back into increasingly sophisticated environmental models which are both used to understand the environment and to plan future trajectories of in-situ assets \cite{Binney2010,fossum2018}. Such approaches, however, are only suitable for co-robotic operation when the quantity of interest is a slowly varying low-dimensional field.


Our proposed approach to co-robotic exploration is as follows. The robot, unsupervised and equipped with a multitude of sensors, constructs a scene model to describe its environment in a concise but meaningful way. This scene model can be tuned to fit the available bandwidth budget, and describe the world at the level of abstraction suitable for describing the phenomena of interest. This scene model is constantly updated with new observation data, and transmitted regularly to the scientist. Given the scene model, the scientist then defines a utility function over various types of scene labels and other high level constraints, 
 which is then broadcast to one or more underwater robots. In the case of multiple robots, it is possible to learn a shared representation through regular message passing between them \cite{Doherty2018}.

The underwater robots can then use the utility function to plan an informative path plan, using either a greedy approach \cite{Girdhar2015a} or more spatially aware planning \cite{Binney2013,Hollinger2013}, collecting more information that is then used to update the scene model. 

\begin{figure}[t]
    \centering
    \includegraphics[width=0.8\linewidth]{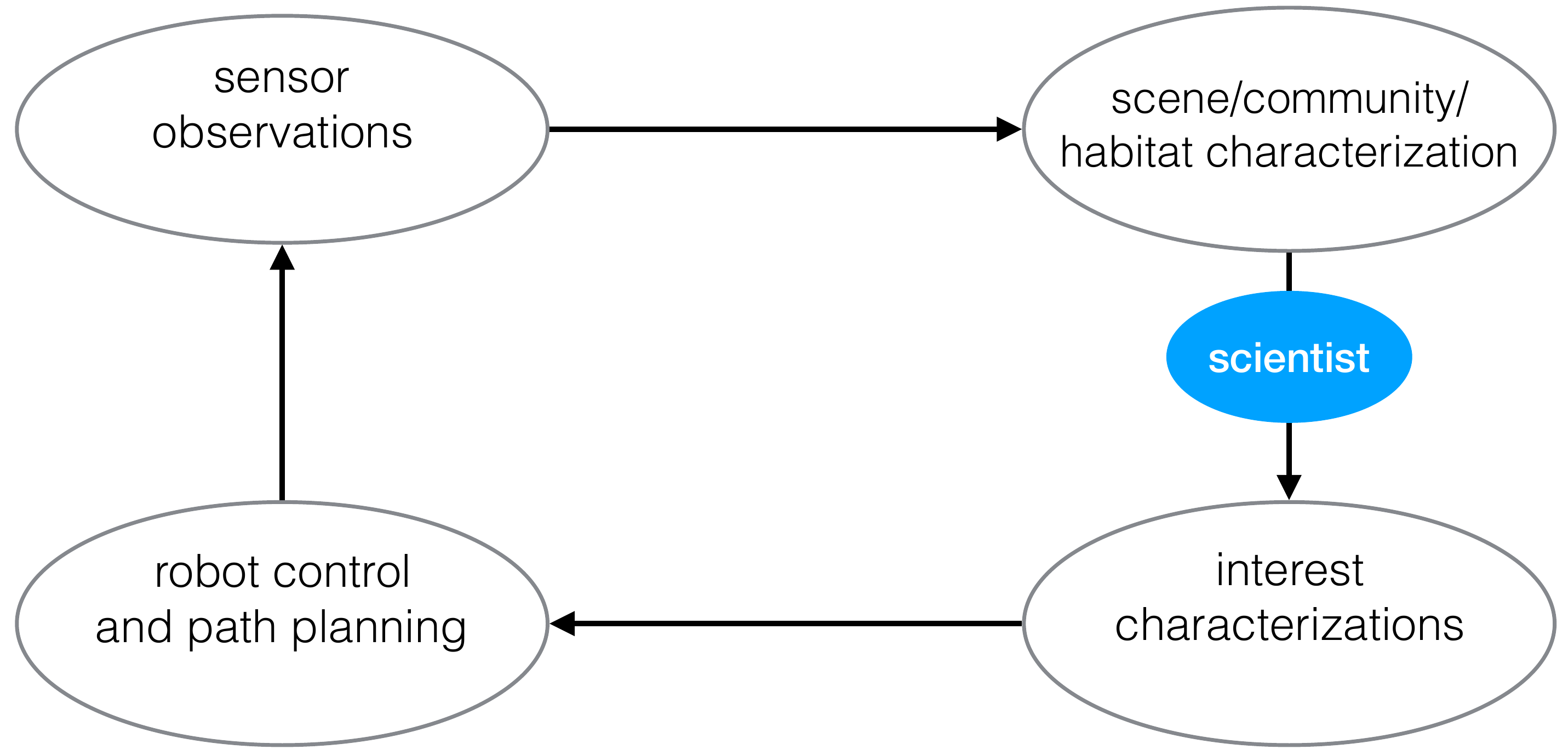}
    \caption{Scientist in the loop, co-robotic exploration. The robot or autonomous agent starts with a set of scientific interest characterizations (lower right), and plans an initial path to maximize mission utility with respect to these interests. After collecting sensor observations and refining a model of the scientific phenomenon, the robot communicates this model to a scientist in-the-loop, who can dictate a set of updated scientific interest characterizations. }
\label{fig:loop}
\end{figure}

This entire process continues in a loop (Fig. \ref{fig:loop}), enabling scientists to continually define high level goals for the robots, even in unknown environments, while letting robots perform low level path planning and sensing autonomously. 


The focus of this paper is on 1) demonstrating a bandwidth-tunable scene understanding technique that can operate in unknown environments and be made to describe the scene at different levels of abstraction by changing only a handful of parameters; and 2) developing the hardware and software infrastructure to enable such an underwater co-robotic exploration mission.

\section{Scene Modeling}

In the context of co-robotic exploration, we require a scene model that can operate in geographic coordinate space, and produce high-level and concise scene descriptions, which are transmitted over a low-bandwidth communication channel such as underwater acoustic communications. In this work we extend HDP-ROST \cite{Girdhar2015} to operate on a stream of image observation data with 3D geographic coordinates. 

\subsection{Generative Process}
The generative model for the observed data is described as follows. At time $t$, we consider generating a set of categorical observations. Every one of these observations $w$ has an associated 3D observed position $(x, y, d) \in \mathbb{R}^3$ that is noisy, and modeled as a random sample from a Gaussian centered around the true position $(x', y', d')$ with some measurement noise $\Sigma$:
$$(x,y,d) \sim N( (x', y', d'), \Sigma),$$
where $d$ is depth. Additionally, each visual observation is associated with a latent scene label $z$, drawn from a location-specific mixture over scene labels $\theta_t$:
$$\theta_t \sim \text{CRP}_{x,y,d}(\gamma, \alpha),$$
$$z \sim \text{Categorical}(\theta_t).$$
We model the distribution of scene labels $z$ using a spatially correlated CRP. We discretize the world into cells, where each cell is represented by one pixel of the transmitted scene map.  Each cell has a Chinese Restaurant (topic model) with an infinite number of tables, each corresponding to different topic labels. We assign a new customer (observation) to one of the $K$ occupied tables with probability proportional to $(n+\alpha)$, where $n$ is the sum of the number of customers sitting at the table in that restaurant and at corresponding tables in the neighboring restaurants, defined using the Von Neumann neighborhood of the cell.  The hyperparameter $\alpha$ prioritizes labels used elsewhere in the scene. The customer sits at a new table with probability proportional to $\gamma$. Thus, the scene label $z$ will take a value $k \in [1,K+1]$, where $K$ is the number of distinct scene labels that we have observed thus far. The advantage of using a CRP is that we do not need to explicitly specify the number of scene labels \textit{a priori}, and it is allowed to grow automatically with the size and complexity of the observation data.

\begin{figure*}[t]
    \centering
    \includegraphics[width=0.24\linewidth]{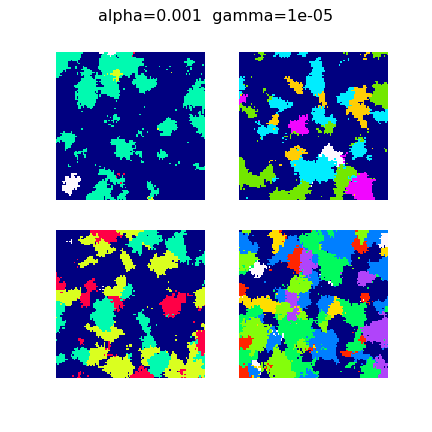}
    \includegraphics[width=0.24\linewidth]{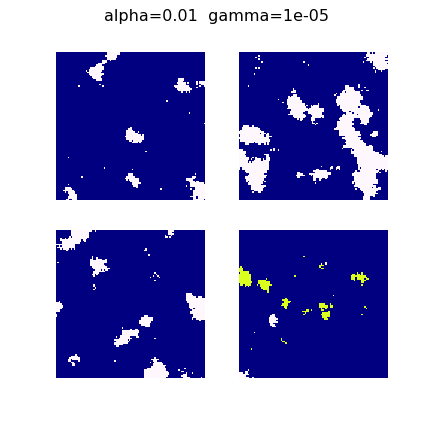}
    \includegraphics[width=0.24\linewidth]{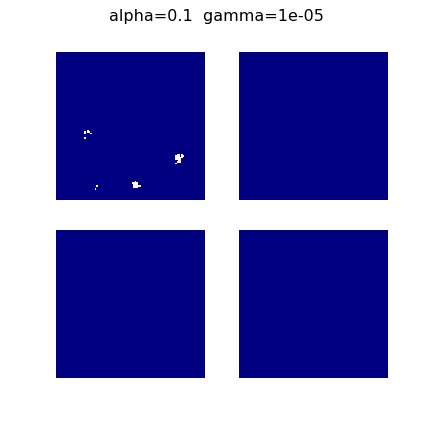}\\

    \includegraphics[width=0.24\linewidth]{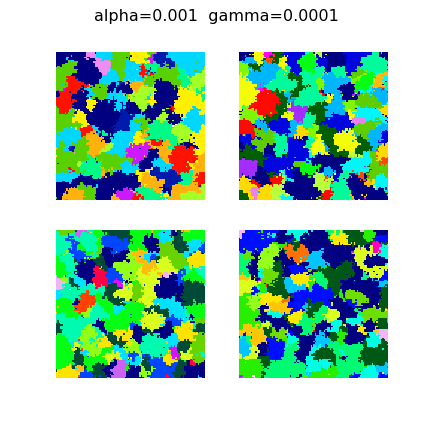}
    \includegraphics[width=0.24\linewidth]{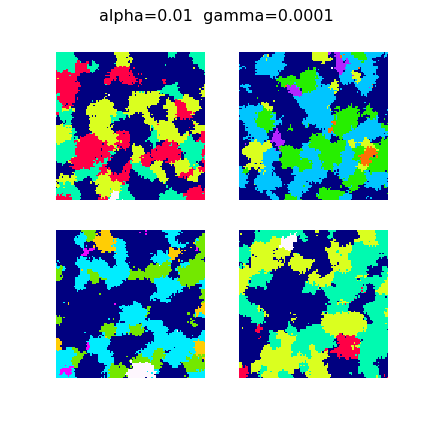}
    \includegraphics[width=0.24\linewidth]{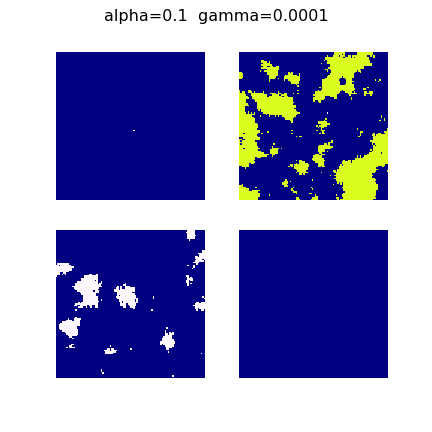}\\

    \includegraphics[width=0.24\linewidth]{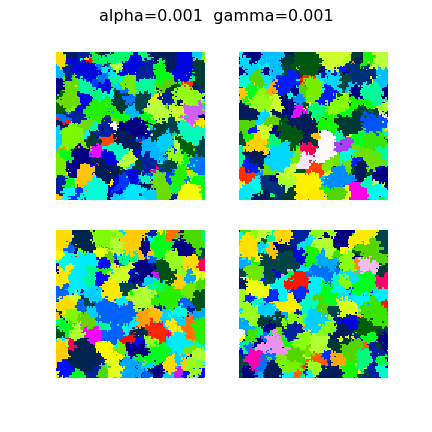}
    \includegraphics[width=0.24\linewidth]{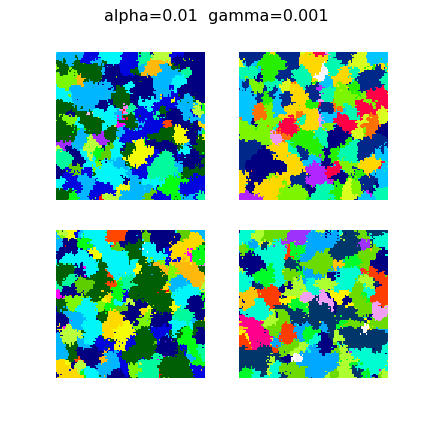}
    \includegraphics[width=0.24\linewidth]{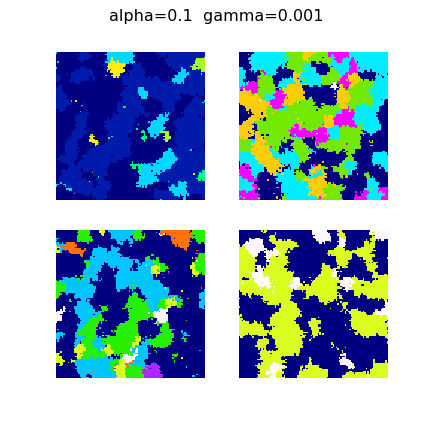}\\
    
    \caption{Examples of randomly generated scenes by the proposed spatially correlated Chinese Restaurant Process. The hyperparameters $\gamma$ and $\alpha$ can be used to tune the generative model for scene with varying complexity in terms of number of different labels and patches. }
\label{fig:generatedworld}
\end{figure*}

Some random samples from the above generative process are shown in Fig.~\ref{fig:generatedworld}. We see that varying $\alpha$ and $\gamma$ gives us random maps with different spatial characteristics in terms of number of different types of patches, and size of the patches. 

Finally, each scene label (topic) is associated with image descriptors (words), which we model using a categorical distribution over a set of $V$ features with the symmetric Dirichlet prior:
$$\phi_k \sim \text{Dirichlet}(\beta),$$
$$w \sim \text{Categorical}(\phi_k).$$
The use of Dirichlet priors biases the generative model towards learning sparse distribution of features used to describe a scene construct.

\subsection{Hyperparameter Selection}
The above generative model can be used to characterize a scene at different levels of abstraction. For example, for a given set of hyperparameters it is possible that we learn labels that correspond to different parts of a coral, or individual coral species, or a colony of different types of corals that often co-exist. On a scientific exploration mission we would like the robot to characterize the scene at the same level of abstraction that is of interest to the scientist. Our proposed solution to enable this behavior is to use an annotated dataset from a location similar to what will be explored, with labels that describe the scene at the desired level of abstraction. Given these annotations $A$, we then find the hyperparameters $\pi = (\alpha, \beta, \gamma)$ for the generative model that maximizes the mutual information $\bm{I}$ between the latent scene labels $Z=\{z_i\}$ and the corresponding annotations $A=\{a_i\}$ using grid search. Assuming that $z = k\in [1,K]$ and $a= j \in [1,J]$:
\begin{eqnarray*}
\pi^{\ast} =(\alpha^{\ast}, \beta^{\ast}, \gamma^{\ast}) &=& \argmax_\pi \bm{I}(Z_\pi, A) \\
\bm{I}(Z_\pi, A) &=& \Sigma_{k}^K \Sigma_{j}^J P(k,j) \log\frac{P(k,j)}{P(k)P(j)}.
\end{eqnarray*}

Note that compared with the standard approach for supervised learning, where the annotations are used to learn all the parameters of the model, we propose to use the annotations only for identifying 3 hyperparameters, while still learning the other parameters of the model in an unsupervised manner. Hence the scene model can still be used to characterize completely unknown environments.

\section{System Design}

\begin{figure}[t]
    \centering
    \includegraphics[width=0.7\linewidth]{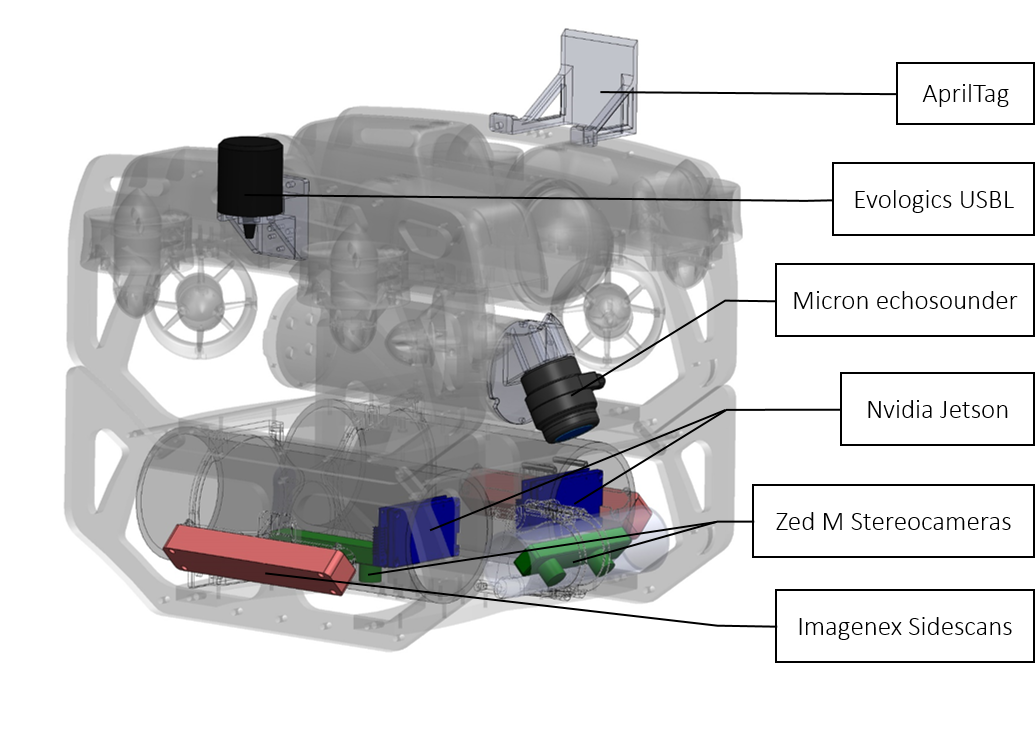}\\
    \includegraphics[width=0.45\linewidth]{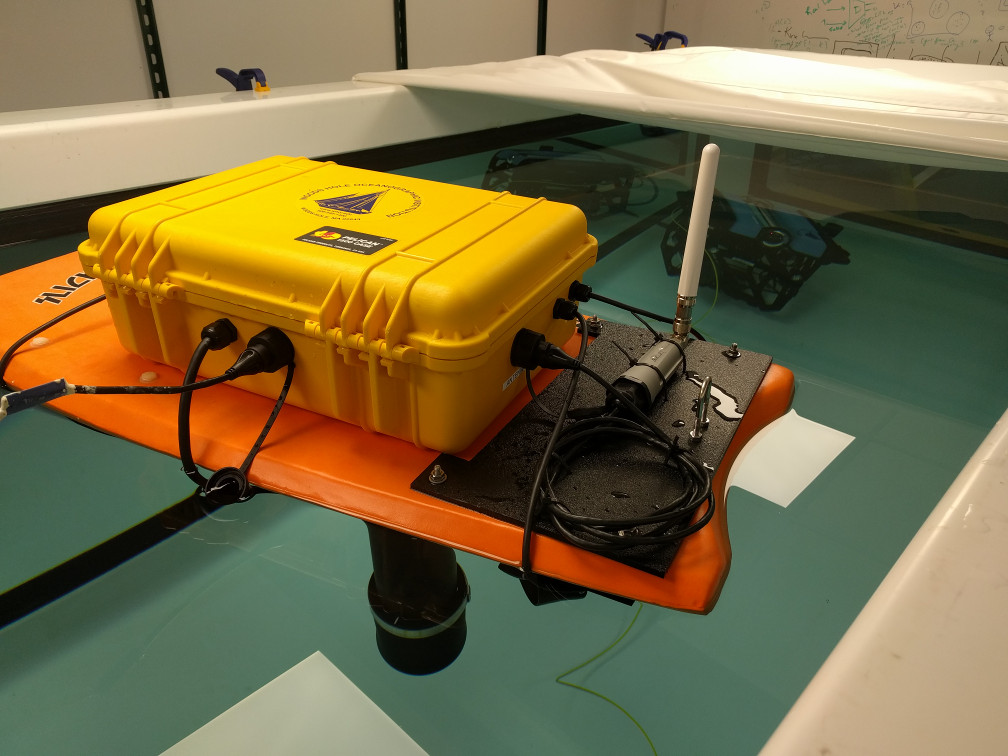}
    \includegraphics[width=0.45\linewidth]{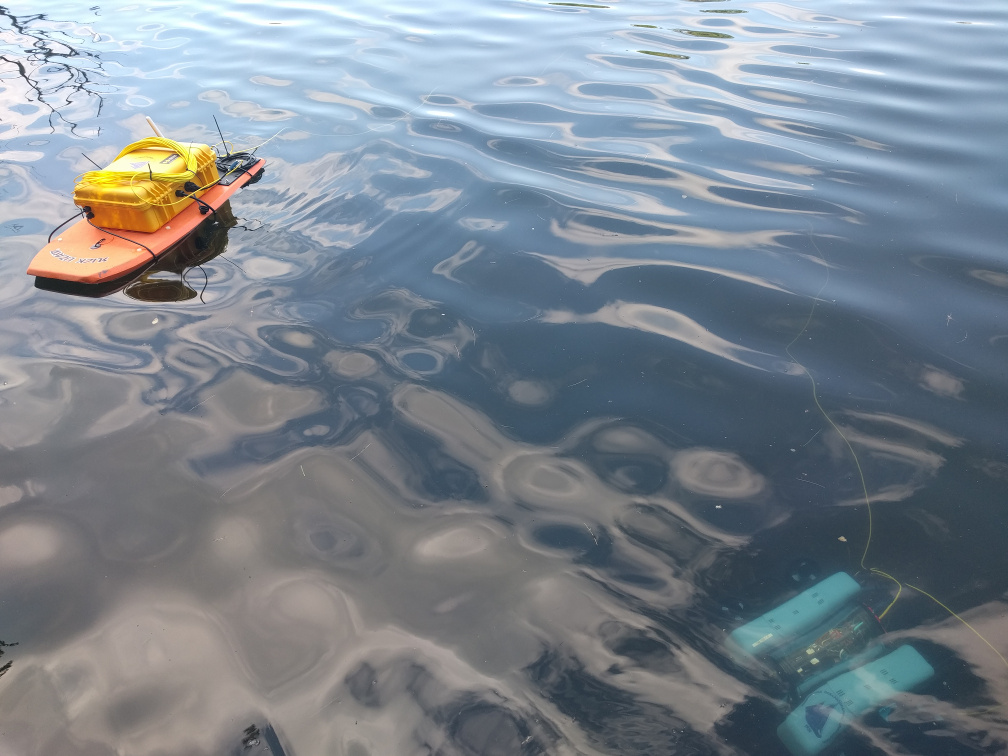}
    \caption{The proposed co-robotic exploration system consists of an autonomous surface vehicle responsible for localizing the AUVs and acting as the communication relay; and one or more AUVs equipped with a multitude of sensors and on-board processing power to enable in-situ scene understanding. Top: The AUV is equipped with bandwidth heavy sensors such as stereo cameras and side scan sonars, in addition to, an echosounder (altitude sensing), acoustic modem, and three computers, two of which have GPUs for computing scene maps in realtime. Bottom Left: The topside vehicle has radios for communicating with the scientists, and a USBL array for acoustic communications and robot localization. Bottom Right: Both the AUV and the ASV deployed in the field.}
\label{fig:hardware}
\end{figure}

Our system design is motivated by the vision that a team of underwater and surface robots with various sensing payloads would be interactively controlled by a scientist in an unknown environment to characterize phenomena observable by the sensing payload. Our current robot system (Fig.~\ref{fig:hardware}) is a step towards this vision, and consists of 1) a surface platform that is used as an air-to-sea communication and localization system to link the scientist to the robots; 2) an AUV equipped with acoustic and optical imaging capabilities, and sufficient computing power to generate concise scene maps from complex high dimensional sensor data streams; 3) a Robot Operating system (ROS) based software architecture to enable a compute graph that spans multiple computers, and optionally multiple robots. 

\subsection{Surface Vehicle}
 The autonomous surface vehicle (ASV) is primarily responsible for acting as a communication and localization relay for the AUV. It is equipped with an Ultra Short Baseline (USBL) acoustic positioning and communication system (shown in Fig. \ref{fig:hardware} bottom-right) to communicate with the underwater robot, and a high speed RF radio to communicate with the scientist on-board the ship or on land. The surface vehicle estimates its distance to the AUV, and fuses a GPS location with the USBL estimate to provide 3D global localization information to the AUV. Furthermore, the surface vehicle is capable of autonomously following the underwater robot to stay directly above it, maximizing bandwidth and localization accuracy. Peak throughput of the acoustic communication system is about 10kbit/sec over 1km range, using the 18-34KHz band. 

\subsection{Underwater Vehicle}
The underwater vehicle (Fig.~\ref{fig:hardware}) is based on the BlueROV2 platform\footnote{The base platform details are at http://docs.bluerobotics.com/brov2/}, with the addition of stereo cameras and side scan sonars for high resolution optical and acoustic imaging, an echosounder for altitude measurement, an acoustic modem/USBL transponder, and two Jetson TX2 computers for on-board processing of sensor data stream. For in-tank localization and testing, AprilTags \cite{Wang2016} are employed. The BlueROV2 has open source hardware and software, and is low cost, which makes it suitable for scaling the robot system to multiple robots, and for easily adding new sensing capabilities. 

\subsection{Software Architecture}

The unsupervised approach to exploration relies upon online robot-side scene modelling to construct a low-bandwidth representation of its surroundings. This representation takes the form of a time-varying scene map constructed from mapping 3D geographic coordinates to their most likely scene label. This can be visualized by the scientist as a colored 3D point cloud, with an injective mapping between colors and scene labels.

\SetKwInOut{Parameter}{Parameter}
\begin{algorithm}
    \label{alg.constructmap}
	\SetAlgoLined
	\SetKwFunction{ExtractFeatures}{ExtractFeatures}
	\SetKwFunction{ComputePoint}{Compute3DFeaturePoint}
	\SetKwFunction{TransformFrame}{TransformFrame}
	\SetKwFunction{InsertObservation}{InsertObservation}
	\SetKwFunction{NotEmpty}{NotEmpty}
	\SetKwFunction{MAPEstimate}{ComputeMAPTopicLabels}
	\SetKwFunction{Mode}{Mode}
	\SetKwFunction{LabelMapLocation}{LabelMap}
	\SetKwData{cell}{cell}
	
	\KwIn{RGB-D image $Im$}
	\KwIn{ROST Topic Model $Tm$}
	\KwIn{Camera-to-world transformation $\mathbf{T}$}
	\Parameter{Camera calibration $\mathbf{K}$}
	\KwResult{Scene map $M$}
	\BlankLine
	$\mathbf{F}\leftarrow$ \ExtractFeatures{$Im$}\\
	\ForEach{$\mathbf{f} \in \mathbf{F}$}{
		$\mathbf{p} \leftarrow$ \ComputePoint{$Im$, $\mathbf{K}$, $\mathbf{f}$}\\
		$\mathbf{f}.xyz \leftarrow$ \TransformFrame{$\mathbf{p}$, $\mathbf{T}$}\\
		\InsertObservation{$Tm$, $\mathbf{f}$}
	}
	\BlankLine
	\ForEach{\cell $\in Tm$}{
		$\mathbf{L} \leftarrow$ \MAPEstimate{$\cell.\mathbf{F}$}\\
		\LabelMapLocation{$M$, $\cell.xyz$, \Mode{$\mathbf{L}$}}
	}
	\caption{Scene Map Construction}
\end{algorithm}

The process of constructing the scene map from the depth image produced by the stereo cameras is given in Algorithm~\ref{alg.constructmap}, and implemented using ROS nodes. The scene map is defined at the resolution of cells, which are sparsely-packed N-dimensional boxes of uniform size, saved in a spatial hash map. Since new cells are added continuously, memory efficient N-D trees are not suitable as they require rebuild after each insertion,  compared to a spatial hash map with $O(1)$ insertion and access time. The cell size is set at initialization to the minimum resolution required to resolve the phenomena of interest.

Since the cell dimensions are upper bounded by the scale of the smallest phenomena of interest, relatively large expanses of uniform terrain (such as sand bars or rocky outcrops) over which the robot may travel will need to be covered by many cells assigned the same topic. This presents an opportunity to reduce the size of the map through compression. Our approach to compression requires first discarding the z-component of the location of each cell, so that the map can be flattened into two dimensions. Then, a 2D map image is created such that each pixel represents the area of a single flattened cell and the image represents the entire region that has been explored. The value of each pixel in the image is set to the positive integer representing the corresponding cell's topic label, or 0 if the cell is empty. Finally, this image is losslessly compressed using the portable network graphic \cite{Duce2003} algorithm to minimize bandwidth, and then transmitted.

Having been presented with the scene map, the operator can learn what a label represents by retrieving representative imagery for that label. This capability will eventually be used to allow the operator to evaluate the relative importance of a label so that the underwater robot can plan an trajectory that maximizes the operator's data collection objectives. 

\begin{figure*}[t]
    \centering
    \includegraphics[width=0.9\linewidth]{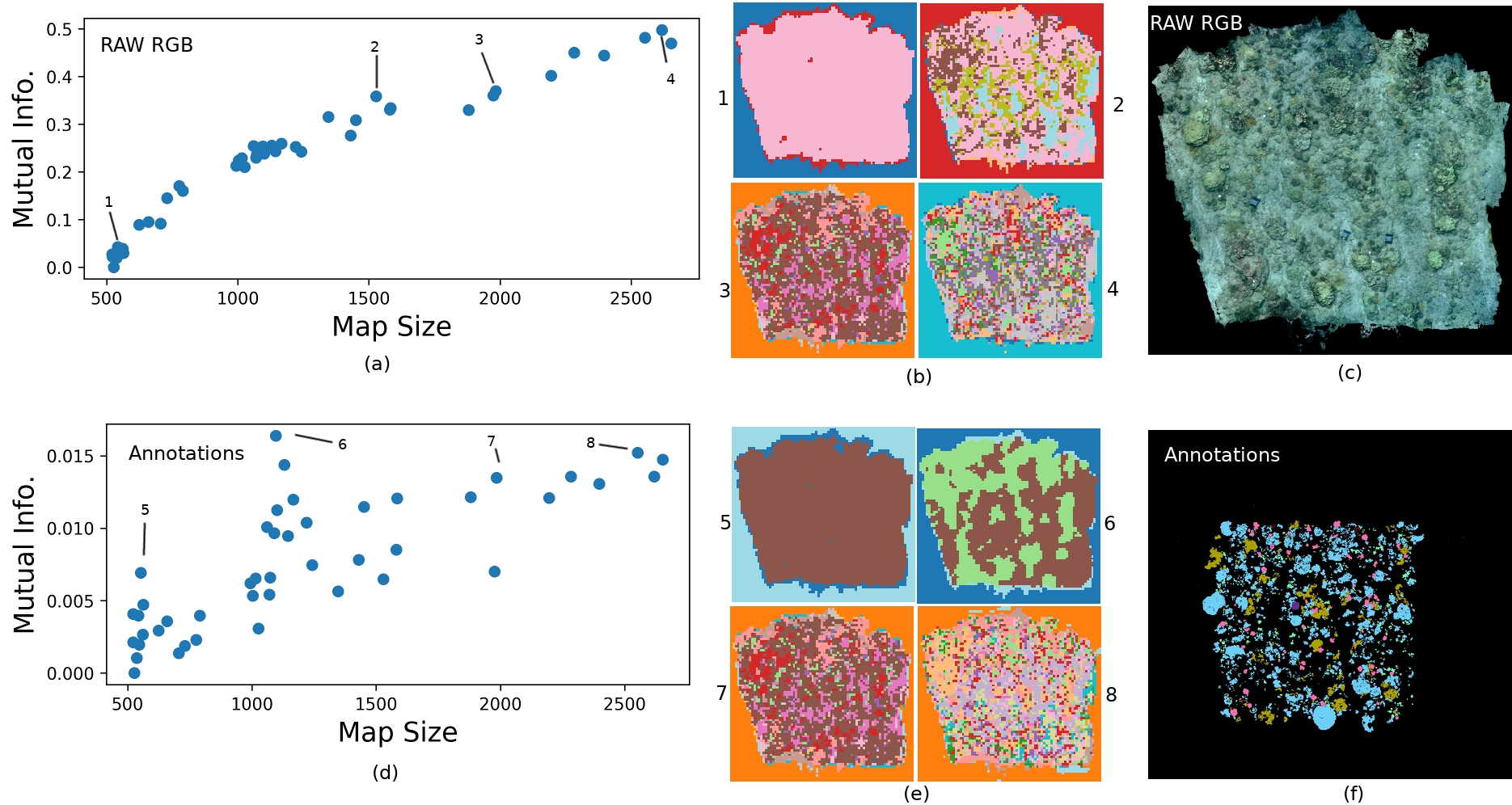}
    \caption{Topic Maps for the HAW-2016-48 Coral Reef Dataset. (a) Scatter plot evaluating different scene maps generated by varying hyperparameters $\alpha, \beta, \gamma$, evaluated on Mutual Information Score with RGB photomosaic map (c) of the reef, and the size of the compressed map in bytes. (d) Scatter plot where the same maps are evaluated by their MI score with the expert annotations (f). MI score is only computed for the region of the reef for which there were annotations. (b,e) show examples of generated map along with their locations in the scatter plots. Variation in the colors of the scene map is purely random, as is only used to distinguish a region from other types of regions.}
\label{fig:coral_results}
\end{figure*}
\begin{figure*}[t]
    \centering

    \includegraphics[width=0.55\linewidth]{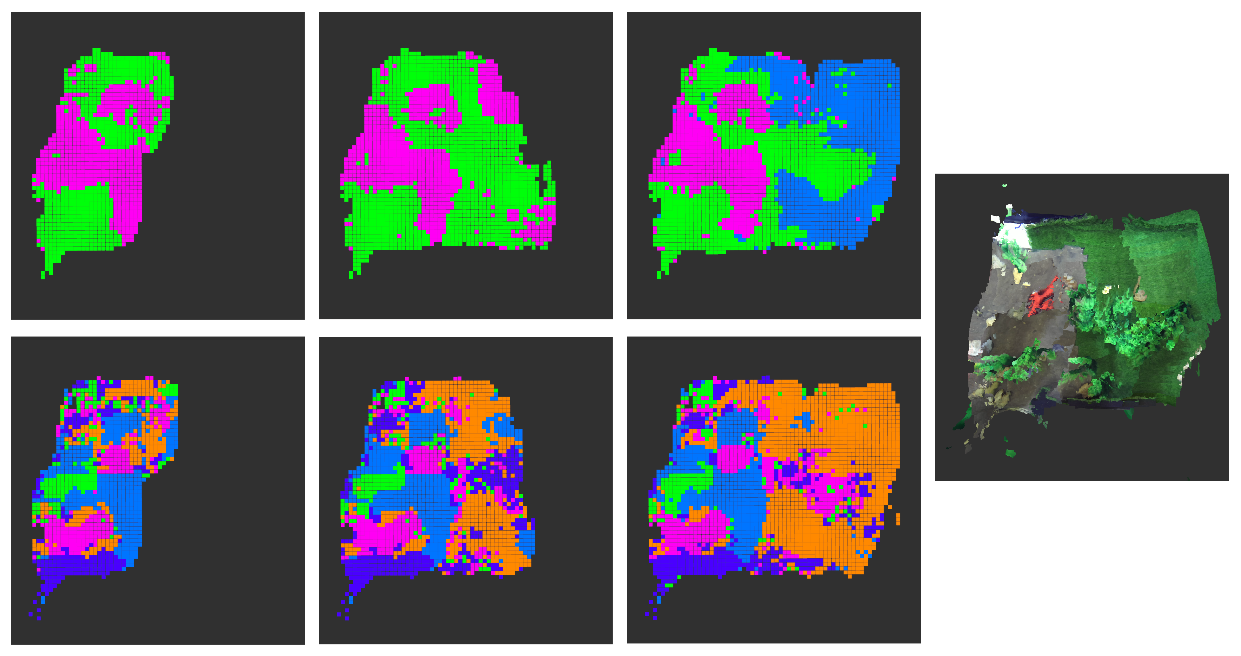}
    \caption{Progression of scene map generated by the robot as it explores an artificially created scene in a test tank, as shown in Fig. \ref{fig:tankrobot}. Top maps correspond to hyperparameters $\alpha = 0.5, \beta = 0.1, \gamma = 0.00001$, bottom maps correspond to $\alpha = 0.00005, \beta = 0.0001, \gamma = 0.0001$. The image on the right shows the stitched map generated using a noisy localization system based on April tags. The maps show that different types of substrates are well characterized by the scene mode even in the presence of noisy location data. }
\label{fig:tank_results}
\end{figure*}

\section{Experiments and Demonstrations}
We evaluate the proposed scene understanding system, as outlined in Algorithm \ref{alg.constructmap}, using a coral reef dataset and then demonstrate the scene modeling system running on our underwater robot in a controlled tank environment with an artificially created scene. In both cases, the robot starts with no prior knowledge of the environment.  

\subsection{Simulated Coral Reef Exploration}
We evaluated the ability of the proposed scene modeling approach to characterize a coral reef environment, given the bandwidth constraints. We used the HAW-2016-48 dataset~\cite{Smith2016} from the 100 Island Challenge\footnote{https://scripps.ucsd.edu/labs/sandin/research/100-island-challenge/} project to simulate observations

Figure \ref{fig:coral_results}(a) shows the mutual information with the directly observed imagery (Fig. \ref{fig:coral_results}(c)) and size of the generated scene map as we vary the scene model hyperparameters $\alpha=\{1.0, 0.1, 0.01, 0.001\}, \beta=\{10.0, 1.0, 0.1\}, \gamma=\{10^{-3}, 10^{-4}, 10^{-5}\}$. Figure \ref{fig:coral_results}(b) shows the same scene maps, but with mutual information computed w.r.t. coral reef species annotation (Fig. \ref{fig:coral_results}(f)).

\subsection{Tank Demonstration}
We demonstrated the scene modelling approach in a physical controlled environment that contained sufficient visual cues, such as artificial seaweed, stones, sand, grass, and a crab, and attempted to build a global topic map from streamed data.

The robot was equipped with outward-facing AprilTags, and an external camera was mounted facing into the tank to provide global localization information that mimicked the position information provided by the surface vehicle via GPS and USBL. The AUV was then autonomously commanded to follow a lawn-mower pattern in the tank. The downward facing Zed stereo camera provided an RGB-labelled point cloud which were then streamed into the topic modelling system. Old data points are updated with new topic information even as new data points are processed, as shown in Fig. \ref{fig:tank_results}.


\section{Conclusion}
We have proposed a novel approach to co-robotic exploration in low bandwidth environments that enables 3D or 2D vision-based robotic exploration while keeping a scientist in the loop. Our approach uses a nonparametric Bayesian scene model to characterize the explored regions of the space in terms of high level, scientifically relevant scene constructs (such as substrate type and coral species), which are then transmitted to the scientist in the form of a scene map. Such scene maps can be used to concisely define utility functions that could be used by robots to perform scientifically relevant informative path planning. Preliminary experiments have demonstrated the ability of the proposed approach to build scene maps that can be transmitted with very low bandwidth requirements, such as while using acoustic communications underwater. We also presented the design of a low cost underwater co-exploration robotic system, and demonstrated the proposed approach on this platform.


\bibliographystyle{IEEEtran}

\bibliography{IEEEabrv,girdhar,shared,claus}

\begin{thebibliography}{10}
\def\url#1{}
\csname url@rmstyle\endcsname
\providecommand{\newblock}{\relax}
\providecommand{\bibinfo}[2]{#2}
\providecommand\BIBentrySTDinterwordspacing{\spaceskip=0pt\relax}
\providecommand\BIBentryALTinterwordstretchfactor{4}
\providecommand\BIBentryALTinterwordspacing{\spaceskip=\fontdimen2\font plus
\BIBentryALTinterwordstretchfactor\fontdimen3\font minus
  \fontdimen4\font\relax}
\providecommand\BIBforeignlanguage[2]{{%
\expandafter\ifx\csname l@#1\endcsname\relax
\typeout{** WARNING: IEEEtran.bst: No hyphenation pattern has been}%
\typeout{** loaded for the language `#1'. Using the pattern for}%
\typeout{** the default language instead.}%
\else
\language=\csname l@#1\endcsname
\fi
#2}}

\bibitem{barker2016}
L.~D.~L. Barker and L.~L. Whitcomb, ``A preliminary survey of underwater
  robotic vehicle design and navigation for under-ice operations,'' in
  \emph{2016 IEEE/RSJ International Conference on Intelligent Robots and
  Systems (IROS)}, Oct 2016, pp. 2028--2035.

\bibitem{cressey2015ocean}
D.~Cressey \emph{et~al.}, ``Ocean-diving robot will not be replaced,'' 2015.

\bibitem{tardioli2015}
\BIBentryALTinterwordspacing
D.~Tardioli, D.~Sicignano, L.~Riazuelo, A.~Romeo, J.~L. Villarroel, and
  L.~Montano, ``Robot teams for intervention in confined and structured
  environments,'' \emph{Journal of Field Robotics}, vol.~33, no.~6, pp.
  765--801, 2015.
  \url{https://onlinelibrary.wiley.com/doi/abs/10.1002/rob.21577}
\BIBentrySTDinterwordspacing

\bibitem{gao2017}
\BIBentryALTinterwordspacing
Y.~Gao and S.~Chien, ``Review on space robotics: Toward top-level science
  through space exploration,'' \emph{Science Robotics}, vol.~2, no.~7, 2017.
  \url{http://robotics.sciencemag.org/content/2/7/eaan5074}
\BIBentrySTDinterwordspacing

\bibitem{nagatani2013}
\BIBentryALTinterwordspacing
K.~Nagatani, S.~Kiribayashi, Y.~Okada, K.~Otake, K.~Yoshida, S.~Tadokoro,
  T.~Nishimura, T.~Yoshida, E.~Koyanagi, M.~Fukushima, and S.~Kawatsuma,
  ``Emergency response to the nuclear accident at the fukushima daiichi nuclear
  power plants using mobile rescue robots,'' \emph{Journal of Field Robotics},
  vol.~30, no.~1, pp. 44--63, 2013.
  \url{https://onlinelibrary.wiley.com/doi/abs/10.1002/rob.21439}
\BIBentrySTDinterwordspacing

\bibitem{yuan2017aerial}
C.~Yuan, Z.~Liu, and Y.~Zhang, ``Aerial images-based forest fire detection for
  firefighting using optical remote sensing techniques and unmanned aerial
  vehicles,'' \emph{Journal of Intelligent \& Robotic Systems}, vol.~88, no.
  2-4, pp. 635--654, 2017.

\bibitem{Teh2010}
Y.~W. Teh and M.~I. Jordan, ``{Hierarchical Bayesian Nonparametric Models with
  Applications},'' \emph{Bayesian nonparametrics}, pp. 158--207, 2010.

\bibitem{Singheaan4809}
\BIBentryALTinterwordspacing
H.~Singh, T.~Maksym, J.~Wilkinson, and G.~Williams, ``Inexpensive, small auvs
  for studying ice-covered polar environments,'' \emph{Science Robotics},
  vol.~2, no.~7, 2017.
  \url{http://robotics.sciencemag.org/content/2/7/eaan4809}
\BIBentrySTDinterwordspacing

\bibitem{Thrun2004}
S.~Thrun, S.~Thayer, W.~Whittaker, C.~Baker, W.~Burgard, D.~Ferguson,
  D.~Hannel, M.~Montemerlo, A.~Morris, Z.~Omohundro, and C.~Reverte,
  ``{Autonomous exploration and mapping of abandoned mines},'' \emph{IEEE
  Robotics {\&} Automation Magazine}, vol.~11, no.~4, pp. 79--91, 12 2004.

\bibitem{weidner2017underwater}
N.~Weidner, S.~Rahman, A.~Q. Li, and I.~Rekleitis, ``Underwater cave mapping
  using stereo vision,'' in \emph{Robotics and Automation (ICRA), 2017 IEEE
  International Conference on}.\hskip 1em plus 0.5em minus 0.4em\relax IEEE,
  2017, pp. 5709--5715.

\bibitem{kinsey2006survey}
J.~C. Kinsey, R.~M. Eustice, and L.~L. Whitcomb, ``A survey of underwater
  vehicle navigation: Recent advances and new challenges,'' in \emph{IFAC
  Conference of Manoeuvering and Control of Marine Craft}, vol.~88, 2006, pp.
  1--12.

\bibitem{Rasmussen:2006:GP}
C.~E. Rasmussen and C.~K.~I. Williams, \emph{{Gaussian Processes for Machine
  Learning}}.\hskip 1em plus 0.5em minus 0.4em\relax The MIT Press, 2006.

\bibitem{Jakuba2011}
\BIBentryALTinterwordspacing
M.~V. Jakuba, D.~Steinberg, J.~C. Kinsey, D.~R. Yoerger, R.~Camilli,
  O.~Pizarro, and S.~B. Williams, ``{Toward automatic classification of
  chemical sensor data from autonomous underwater vehicles},'' in \emph{2011
  IEEE/RSJ International Conference on Intelligent Robots and Systems}.\hskip
  1em plus 0.5em minus 0.4em\relax IEEE, 9 2011, pp. 4722--4727.
  \url{http://ieeexplore.ieee.org/lpdocs/epic03/wrapper.htm?arnumber=6095158}
\BIBentrySTDinterwordspacing

\bibitem{Das2015}
\BIBentryALTinterwordspacing
J.~Das, F.~Py, J.~B.~J. Harvey, J.~P. Ryan, A.~Gellene, R.~Graham, D.~A. Caron,
  K.~Rajan, and G.~S. Sukhatme, ``{Data-driven robotic sampling for marine
  ecosystem monitoring},'' \emph{The International Journal of Robotics
  Research}, vol.~34, no.~12, pp. 1435--1452, 10 2015.
  \url{http://ijr.sagepub.com/cgi/doi/10.1177/0278364915587723}
\BIBentrySTDinterwordspacing

\bibitem{Arora2017clean}
\BIBentryALTinterwordspacing
A.~Arora, R.~Fitch, and S.~Sukkarieh, ``{An Approach to Autonomous Science by
  Modeling Geological Knowledge in a Bayesian Framework},'' \emph{2017 IEEE/RSJ
  International Conference on Intelligent Robots and Systems (IROS)}, pp.
  3803--3810, 3 2017.  \url{http://arxiv.org/abs/1703.03146}
\BIBentrySTDinterwordspacing

\bibitem{Zhang2016}
\BIBentryALTinterwordspacing
Y.~Zhang, J.~G. Bellingham, J.~P. Ryan, B.~Kieft, and M.~J. Stanway,
  ``{Autonomous Four-Dimensional Mapping and Tracking of a Coastal Upwelling
  Front by an Autonomous Underwater Vehicle},'' \emph{Journal of Field
  Robotics}, vol.~33, no.~1, pp. 67--81, 1 2016.
  \url{http://doi.wiley.com/10.1002/rob.21617}
\BIBentrySTDinterwordspacing

\bibitem{Gautam2017}
\BIBentryALTinterwordspacing
S.~Gautam, B.~S. Roy, A.~Candela, and D.~Wettergreen, ``{Science-aware
  exploration using entropy-based planning},'' in \emph{2017 IEEE/RSJ
  International Conference on Intelligent Robots and Systems (IROS)}.\hskip 1em
  plus 0.5em minus 0.4em\relax IEEE, 9 2017, pp. 3819--3825.
  \url{http://ieeexplore.ieee.org/document/8206232/}
\BIBentrySTDinterwordspacing

\bibitem{Girdhar2015Gibbs}
\BIBentryALTinterwordspacing
Y.~Girdhar and G.~Dudek, ``{Gibbs Sampling Strategies for Semantic Perception
  of Streaming Video Data},'' \emph{ArXiv e-prints}, p.~7, 2015.
  \url{http://arxiv.org/abs/1509.03242}
\BIBentrySTDinterwordspacing

\bibitem{Sudderth2009}
E.~B. Sudderth and M.~I. Jordan, ``{Shared Segmentation of Natural Scenes Using
  Dependent Pitman-Yor Processes},'' in \emph{Advances in Neural Information
  Processing Systems 21 (NIPS 2008)}, 2009.

\bibitem{Joho2012}
D.~Joho, G.~D. Tipaldi, N.~Engelhard, C.~Stachniss, and W.~Burgard,
  ``{Nonparametric Bayesian Models for Unsupervised Scene Analysis and
  Reconstruction},'' \emph{Robotics Science and Systems}, 2012.

\bibitem{steinberg2011bayesian}
D.~Steinberg, A.~Friedman, O.~Pizarro, and S.~B. Williams, ``A bayesian
  nonparametric approach to clustering data from underwater robotic surveys,''
  in \emph{International Symposium on Robotics Research}, vol.~28, 2011, pp.
  1--16.

\bibitem{Girdhar2015}
\BIBentryALTinterwordspacing
Y.~Girdhar, {Walter Cho}, M.~Campbell, J.~Pineda, E.~Clarke, and H.~Singh,
  ``{Anomaly detection in unstructured environments using Bayesian
  nonparametric scene modeling},'' in \emph{2016 IEEE International Conference
  on Robotics and Automation (ICRA)}.\hskip 1em plus 0.5em minus 0.4em\relax
  IEEE, 5 2016, pp. 2651--2656.  \url{http://arxiv.org/abs/1509.07979
  http://ieeexplore.ieee.org/document/7487424/}
\BIBentrySTDinterwordspacing

\bibitem{michel2018}
\BIBentryALTinterwordspacing
A.~P. Michel, S.~D. Wankel, J.~Kapit, Z.~Sandwith, and P.~R. Girguis, ``In situ
  carbon isotopic exploration of an active submarine volcano,'' \emph{Deep Sea
  Research Part II: Topical Studies in Oceanography}, vol. 150, pp. 57 -- 66,
  2018, results of Telepresence-Enabled Oceanographic Exploration.
  \url{http://www.sciencedirect.com/science/article/pii/S0967064517301224}
\BIBentrySTDinterwordspacing

\bibitem{mcveigh2018}
\BIBentryALTinterwordspacing
D.~McVeigh, A.~Skarke, A.~Dekas, C.~Borrelli, W.-L. Hong, J.~Marlow,
  A.~Pasulka, S.~Jungbluth, R.~Barco, and A.~Djurhuus, ``Characterization of
  benthic biogeochemistry and ecology at three methane seep sites on the
  northern u.s. atlantic margin,'' \emph{Deep Sea Research Part II: Topical
  Studies in Oceanography}, vol. 150, pp. 41 -- 56, 2018, results of
  Telepresence-Enabled Oceanographic Exploration.
  \url{http://www.sciencedirect.com/science/article/pii/S0967064517301601}
\BIBentrySTDinterwordspacing

\bibitem{everett2018}
\BIBentryALTinterwordspacing
M.~V. Everett and L.~K. Park, ``Exploring deep-water coral communities using
  environmental dna,'' \emph{Deep Sea Research Part II: Topical Studies in
  Oceanography}, vol. 150, pp. 229 -- 241, 2018, results of
  Telepresence-Enabled Oceanographic Exploration.
  \url{http://www.sciencedirect.com/science/article/pii/S0967064517301546}
\BIBentrySTDinterwordspacing

\bibitem{jakuba2018}
M.~V. Jakuba, C.~R. German, A.~D. Bowen, L.~L. Whitcomb, K.~Hand, A.~Branch,
  S.~Chien, and C.~McFarland, ``Teleoperation and robotics under ice:
  Implications for planetary exploration,'' in \emph{2018 IEEE Aerospace
  Conference}, March 2018, pp. 1--14.

\bibitem{mensi2014}
\BIBentryALTinterwordspacing
B.~Mensi, R.~Rowe, S.~Dees, D.~Bryant, D.~Jones, and R.~Carr, ``{Operational
  glider monitoring, piloting, and communications},'' in \emph{2014 IEEE/OES
  Autonomous Underwater Vehicles (AUV)}.\hskip 1em plus 0.5em minus 0.4em\relax
  IEEE, 10 2014, pp. 1--3.  \url{http://ieeexplore.ieee.org/document/7054415/}
\BIBentrySTDinterwordspacing

\bibitem{flexas2018}
M.~M. Flexas, M.~I. Troesch, S.~Chien, A.~F. Thompson, S.~Chu, A.~Branch, J.~D.
  Farrara, and Y.~Chao, ``Autonomous sampling of ocean submesoscale fronts with
  ocean gliders and numerical model forecasting,'' \emph{Journal of Atmospheric
  and Oceanic Technology}, vol. 35 (3), March 2018.

\bibitem{Binney2010}
J.~Binney, A.~Krause, and G.~S. Sukhatme, ``{Informative Path Planning for an
  Autonomous Underwater Vehicle},'' \emph{Robotics and automation (ICRA), 2010
  IEEE international conference on.}, 2010.

\bibitem{fossum2018}
\BIBentryALTinterwordspacing
T.~O. Fossum, J.~Eidsvik, I.~Ellingsen, M.~O. Alver, G.~M. Fragoso, G.~Johnsen,
  R.~Mendes, M.~Ludvigsen, and K.~Rajan, ``Information-driven robotic sampling
  in the coastal ocean,'' \emph{Journal of Field Robotics}, vol.~0, no.~0,
  2018.  \url{https://onlinelibrary.wiley.com/doi/abs/10.1002/rob.21805}
\BIBentrySTDinterwordspacing

\bibitem{Doherty2018}
K.~Doherty, G.~Flaspohler, N.~Roy, and Y.~Girdhar, ``{Approximate Distributed
  Spatiotemporal Topic Models for Multi-Robot Terrain Characterization},'' in
  \emph{Intelligent Robots and Systems (IROS)}, 2018.

\bibitem{Girdhar2015a}
\BIBentryALTinterwordspacing
Y.~Girdhar and G.~Dudek, ``{Modeling curiosity in a mobile robot for long-term
  autonomous exploration and monitoring},'' \emph{Autonomous Robots}, vol.~40,
  no.~7, pp. 1267--1278, 10 2016.
  \url{http://link.springer.com/10.1007/s10514-015-9500-x}
\BIBentrySTDinterwordspacing

\bibitem{Binney2013}
\BIBentryALTinterwordspacing
J.~Binney, A.~Krause, and G.~S. Sukhatme, ``{Optimizing waypoints for
  monitoring spatiotemporal phenomena},'' \emph{The International Journal of
  Robotics Research}, vol.~32, no.~8, pp. 873--888, 7 2013.
  \url{http://ijr.sagepub.com/cgi/doi/10.1177/0278364913488427}
\BIBentrySTDinterwordspacing

\bibitem{Hollinger2013}
\BIBentryALTinterwordspacing
G.~A. Hollinger and G.~S. Sukhatme, ``{Sampling-based Motion Planning for
  Robotic Information Gathering},'' in \emph{Robotics: Science and Systems},
  2013, pp. 72--983.  \url{http://www.roboticsproceedings.org/rss09/p51.pdf}
\BIBentrySTDinterwordspacing

\bibitem{Wang2016}
J.~Wang and E.~Olson, ``{AprilTag 2: Efficient and robust fiducial
  detection},'' in \emph{IEEE International Conference on Intelligent Robots
  and Systems}, 2016.

\bibitem{Duce2003}
\BIBentryALTinterwordspacing
D.~Duce, M.~Adler, T.~Boutell, J.~Bowler, C.~Brunschen, A.~M. Costello, L.~D.
  Crocker, A.~Dilger, O.~Fromme, J.-l. Gailly, C.~Herborth, A.~Jakulin,
  N.~Kettler, T.~Lane, A.~Lehmann, C.~Lilley, D.~Martindale, O.~Mortensen,
  K.~S. Pickens, R.~P. Poole, G.~Randers-Pehrson, G.~Roelofs, W.~van Schaik,
  G.~Schalnat, P.~Schmidt, M.~Stokes, T.~Wegner, and J.~Wohl, ``{Portable
  Network Graphics (PNG) Specification (Second Edition)},'' W3C, Tech. Rep.,
  2003.  \url{https://www.w3.org/TR/2003/REC-PNG-20031110/{\#}F-Relationship}
\BIBentrySTDinterwordspacing

\bibitem{Smith2016}
\BIBentryALTinterwordspacing
J.~E. Smith, R.~Brainard, A.~Carter, S.~Grillo, C.~Edwards, J.~Harris,
  L.~Lewis, D.~Obura, F.~Rohwer, E.~Sala, P.~S. Vroom, and S.~Sandin,
  ``{Re-evaluating the health of coral reef communities: baselines and evidence
  for human impacts across the central Pacific},'' \emph{Proceedings of the
  Royal Society B: Biological Sciences}, vol. 283, no. 1822, p. 20151985, 1
  2016.
  \url{http://rspb.royalsocietypublishing.org/lookup/doi/10.1098/rspb.2015.1985}
\BIBentrySTDinterwordspacing

\end{thebibliography}

\end{document}